\documentclass[journal]{IEEEtran}
\usepackage{graphicx}
\usepackage{amssymb}
\usepackage{lineno}
\usepackage{hyperref}
\usepackage{amsmath}
\usepackage{algorithm} 
\usepackage{algpseudocode} 
\usepackage{enumerate}
\usepackage{hyperref}
\usepackage{booktabs}
\usepackage{multirow} 
\usepackage{rotating}
\usepackage{pifont}
\usepackage{booktabs} 
\usepackage{tabularx} 
\usepackage{siunitx}  
\usepackage{ragged2e} 
\usepackage{array}    
\usepackage{subcaption}
\usepackage{tabularx} %
\usepackage{caption}  
\usepackage[table,xcdraw]{xcolor}
\usepackage{authblk}
\usepackage{orcidlink}

\usepackage[pscoord]{eso-pic}

\usepackage{amsmath, amssymb, amsthm}
\usepackage{array}
\usepackage{color}
\usepackage{tikz}
\usepackage{textcomp}
\usepackage{hyperref}
\usepackage{lipsum}

\newcolumntype{C}{>{\centering\arraybackslash}X}

\newcommand{\xmark}{\ding{55}}%

\newcommand*{\updatedText}[1]{\textcolor{black}{ #1}}

\begin{document}

\title{Biological Brain Age Estimation using Sex-Aware Adversarial Variational Autoencoder with Multimodal Neuroimages}

\author[1]{Abd~Ur~Rehman\orcidlink{0000-0003-3120-9264}}
\author[2]{Azka~Rehman\orcidlink{0000-0001-7085-0105}}
\author[3,*]{Muhammad~Usman\orcidlink{0000-0002-7278-8488}}
\author[4]{Abdullah~Shahid\orcidlink{0000-0000-0000-0003}}
\author[5]{Sung-Min~Gho}
\author[6]{Aleum~Lee}
\author[7]{Tariq~M.~Khan\orcidlink{0000-0002-7477-1591}}
\author[8]{Imran~Razzak\orcidlink{0000-0002-3930-6600}}

\affil[1]{Dongguk University, Seoul, 04620, South Korea}
\affil[2]{Seoul National University Graduate School, Seoul, 03080, Republic of Korea}
\affil[3]{Stanford University, CA 94305, USA}
\affil[4]{DeepChain AI\&IT Technologies, Islamabad 44000, Pakistan}
\affil[5]{Medical R\&D Center, DeepNoid Inc., Seoul, 08376, South Korea}
\affil[6]{Soonchunhyang University Bucheon Hospital, Bucheon-si, Gyeonggi-do, 14584, South Korea}
\affil[7]{Naif Arab University for Security Sciences, Riyadh, 11452, Kingdom of Saudi Arabia}
\affil[8]{University of New South Wales, Sydney, NSW 2052, Australia}
\affil[*]{Corresponding E-mail: \href{mailto:usmanm@stanford.edu}{usmanm@stanford.edu}}

\maketitle


\begin{abstract}
Brain aging involves structural and functional changes and therefore serves as a key biomarker for brain health. Combining structural magnetic resonance imaging (sMRI) and functional magnetic resonance imaging (fMRI) has the potential to improve brain age estimation by leveraging complementary data. However, fMRI data, being noisier than sMRI, complicates multimodal fusion. Traditional fusion methods often introduce more noise than useful information, which can reduce accuracy compared to using sMRI alone. In this paper, we propose a novel multimodal framework for biological brain age estimation, utilizing a sex-aware adversarial variational autoencoder (SA-AVAE). Our framework integrates adversarial and variational learning to effectively disentangle the latent features from both modalities. Specifically, we decompose the latent space into modality-specific codes and shared codes to represent complementary and common information across modalities, respectively. To enhance the disentanglement, we introduce cross-reconstruction and shared-distinct distance ratio loss as regularization terms. Importantly, we incorporate sex information into the learned latent code, enabling the model to capture sex-specific aging patterns for brain age estimation via an integrated regressor module. We evaluate our model using the publicly available OpenBHB dataset, a comprehensive multi-site dataset for brain age estimation. The results from ablation studies and comparisons with state-of-the-art methods demonstrate that our framework outperforms existing approaches and shows significant robustness across various age groups, highlighting its potential for real-time clinical applications in the early detection of neurodegenerative diseases.
\end{abstract}


\IEEEpeerreviewmaketitle

\section{Introduction}
The aging population poses significant global challenges, profoundly affecting economic, medical, and societal systems \cite{khan2019population}. Among these challenges, declining brain function and neurodegenerative diseases in the elderly exacerbate the burden on healthcare and social structures \cite{reeve2014ageing}. In the fields of life sciences and biomedicine, predicting and assessing age-related neurodegeneration, as well as developing treatments to mitigate or reverse its effects, remain critical research priorities \cite{farooqui2009aging}. A key focus is the distinction between biological brain age and chronological age, which serves as an informative biomarker for neurological disorders such as Parkinson’s disease \cite{beheshti2020t1}, vascular dementia \cite{haan2004can}, mild cognitive impairment (MCI), and Alzheimer’s disease (AD) \cite{franke2012longitudinal}. Deviations from the typical brain aging trajectory are particularly significant, as they can predict an individual's future risk of developing neurodegenerative conditions \cite{steffener2016differences}. Modeling brain aging patterns through neuroimaging data and tracking individual trajectories offers a promising approach to understanding brain aging dynamics and its implications for neurological health \cite{cole2017predicting}.

\updatedText{Neuroimaging techniques, particularly magnetic resonance imaging (MRI), are powerful tools for studying age-related changes in brain structure and function, such as morphometric atrophy and reduced connectivity \cite{cole2019quantification}. Multimodal MRI, which combines structural MRI (sMRI) and functional MRI (fMRI), provides complementary insights by capturing both structural and functional aspects of the living human brain under noninvasive or minimally invasive conditions. This approach has been widely employed in research on brain aging and the clinical diagnosis of neurological diseases \cite{plis2018reading}. However, fMRI presents challenges due to its relatively low spatial resolution, high noise levels, and the immature and rapidly changing nature of functional connectivity derived from it. These limitations make it impractical or ineffective to directly fuse sMRI and fMRI data using conventional multimodal fusion strategies \cite{corps2019morphological}. In some cases, such strategies can even diminish the accuracy of sMRI-derived features, which are well-established as robust biomarkers for biological age estimation. Therefore, effective fusion of sMRI and fMRI data requires approaches that minimize the negative impact of one modality on the other during the fusion process, ensuring that each modality's strengths are preserved and effectively utilized.}

\begin{figure*}[t]
\centering
\includegraphics[width=1\textwidth]{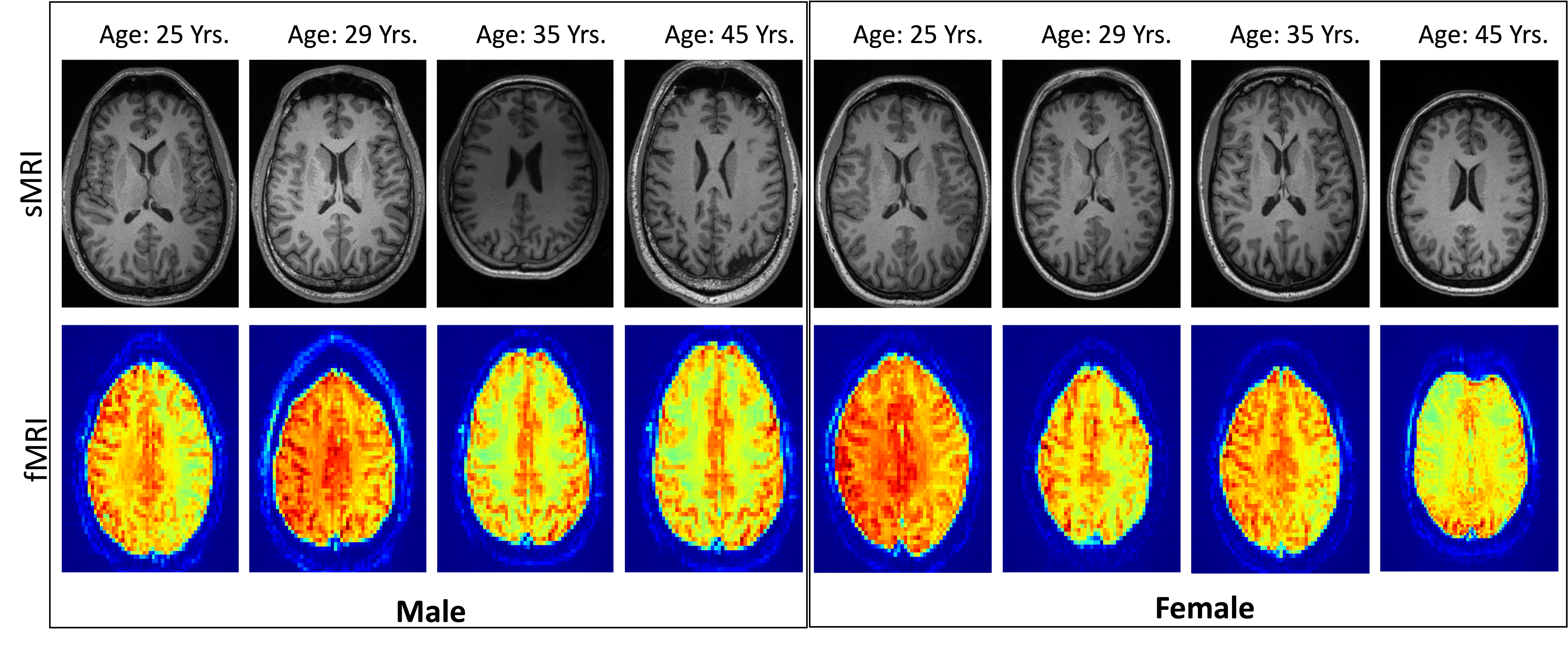}
\caption{\updatedText{Visualization of structural magnetic resonance imaging (sMRI) and functional magnetic resonance imaging (fMRI) in male and female patients across various age groups. sMRI reveals the anatomical details of the brain, whereas fMRI depicts brain activity by measuring changes in blood flow; this is shown on a color scale where warmer colors typically indicate higher levels of activity. }}
\label{samples_visualization}
\end{figure*}

Despite the potential of multimodal neuroimaging, research on its use for brain age estimation, particularly in exploring anatomical and functional differences between male and female brains, has been limited. Our study aims to address this gap by estimating brain age using both fMRI and sMRI data, identifying potential indicators of brain development through various imaging modalities. The inherent differences between sMRI and fMRI, including noise levels, spatial resolution, and the dynamic nature of functional connectivity, highlight the need for innovative fusion techniques that leverage sMRI's precision and reliability as biomarkers for age prediction \cite{bib_10, bib_11}.

\updatedText{To overcome these challenges, it is essential to address the role of sex information in brain age estimation. Figure~\ref{samples_visualization} shows samples of male and female brain sMRI and fMRI scans across different ages. Research indicates significant anatomical and functional differences between male and female brains, which influence the aging process \cite{franke2010estimating, cole2017predicting}. Although sex information has proven effective when incorporated into deep learning models \cite{ wilms2022invertible}, its integration into multimodal neuroimaging studies has been scarce. Our work incorporates sex information into a multimodal framework, enhancing the precision and robustness of brain age estimation.}

\begin{table*}[t]
\centering
\caption{\updatedText{Bird's-eye view comparison of our proposed framework with previously published studies.}}
\color{black}
\label{litrature_tab}
\scalebox{0.9}{
\begin{tabularx}{\textwidth}{@{}lCCCC@{}}
\toprule
\textbf{\footnotesize Study (Year)} & \textbf{\footnotesize Multimodal Learning} & \textbf{\footnotesize Sex Information}  & \textbf{\footnotesize Adversarial learning} & \textbf{\footnotesize Disentangled learning} \\
\midrule

Liem \textit{et al.} 2017 \cite{liem2017predicting} & \checkmark & \xmark & \xmark & \xmark \\


He \textit{et al.} 2021 \cite{he2021multi} & \checkmark & \xmark & \xmark & \xmark \\


Cheng \textit{et al.} 2021 \cite{cheng2021brain} & \xmark & \checkmark & \xmark & \xmark \\

Armanious \textit{et al.} 2021 \cite{armanious2021age} & \xmark & \checkmark & \xmark & \xmark \\



Mouches \textit{et al.} 2022 \cite{mouches2022multimodal} & \checkmark & \xmark & \xmark & \xmark \\

Cai \textit{et al.} 2023 \cite{cai2023graph} & \checkmark & \xmark & \xmark & \checkmark \\

Liu \textit{et al.} 2023 \cite{liu2023risk} & \xmark & \checkmark & \xmark & \xmark \\

Hu \textit{et al.} 2024 \cite{hu2020disentangled} & \checkmark & \checkmark & \xmark & \checkmark \\

\textbf{Our Study} & \checkmark & \checkmark & \checkmark & \checkmark  \\ 
\bottomrule
\end{tabularx} }
\end{table*}

\updatedText{To address these challenges, we introduce a novel multimodal learning framework, the Sex-Aware Adversarial Variational Autoencoder (SA-AVAE). The SA-AVAE framework uniquely integrates adversarial and variational autoencoder losses to disentangle shared and modality-specific features in a latent space, effectively reducing noise and improving feature representation. This disentanglement ensures that the shared latent space captures complementary information across sMRI and fMRI, while the modality-specific latent spaces retain unique characteristics, minimizing interference during fusion. Furthermore, the adversarial component suppresses noise by enforcing the separation of irrelevant features, while the variational component enhances feature disentanglement by learning distinct probabilistic distributions for shared and unique features.}

\updatedText{Additionally, the SA-AVAE framework incorporates sex information into the latent space, leveraging known anatomical and functional differences between male and female brains \cite{franke2010estimating, cole2017predicting}. By integrating sex information, the model not only improves accuracy but also enhances robustness by accounting for gender-related variations. This approach represents a significant advancement over traditional fusion techniques that often overlook such critical demographic factors.}

Following are the key contributions of this study: 

\begin{itemize}
    \item We propose a multimodal framework, named the Sex-Aware Adversarial Variational Autoencoder (SA-AVAE), to estimate brain age using multimodal MRI scans (sMRI and fMRI). This framework is the first to integrate sex information directly into the latent space, enhancing the model's accuracy and robustness.
    \item The proposed framework generates a disentangled latent space by enforcing distinct distributions through the simultaneous application of adversarial and variational autoencoder losses within a unified architecture.
    \item We present a comprehensive strategy for fine-tuning loss weight parameters, providing valuable insights for optimizing architectures in broader applications.
    \item Extensive experimentation on publicly available datasets demonstrates the accuracy and robustness of the SA-AVAE model, setting a new standard for brain age estimation techniques.
\end{itemize}

The remainder of the paper is organized as follows: Section \ref{rw_section} presents a comprehensive review of related studies, providing the context and background for our work. Section \ref{pro} details the proposed framework, including its various components and the underlying principles. Section \ref{es_section} elaborates on the dataset specifications and preprocessing steps, as well as the architectural details and training strategy of the model. Section \ref{res} showcases the experimental results, accompanied by an in-depth analysis. Finally, Section \ref{con} concludes the paper by summarizing the key findings and discussing future research directions, with a focus on potential improvements and extensions to the proposed approach.

\section{Related Work}
\label{rw_section}
\updatedText{The application of deep learning techniques for brain age estimation using MRI has seen significant advancements in recent years. In particular, Convolutional Neural Networks (CNNs) have emerged as the primary architecture for estimating brain age from structural MRI (sMRI) data, offering the ability to learn complex image features directly from the data. Early works in this domain employed 2D CNNs applied to individual slices of sMRI scans \cite{feng2020estimating}, while more recent studies have adopted 3D CNNs, which analyze the entire brain to capture more comprehensive structural features for brain age prediction \cite{he2021multi}. Dinsdale et al. \cite{dinsdale2021learning} introduced a 3D CNN model that leveraged whole-brain MRI data, demonstrating improved accuracy by capturing global structural features related to brain aging.}

\updatedText{However, training deep learning models on small datasets remains a significant challenge. To address this, Peng et al. \cite{peng2021accurate} proposed a 3D fully convolutional network that reduces model complexity by using fewer parameters, thereby improving efficiency without compromising prediction accuracy. Cheng et al. \cite{cheng2021brain} further improved performance by introducing a two-stage age prediction network (TSAN), which first provides a rough estimation followed by a refinement stage. This multi-stage approach allows the model to progressively enhance its representations of brain aging.}

\updatedText{In addition to CNN-based architectures, the incorporation of attention mechanisms has shown promise in improving brain age prediction. He et al. \cite{he2021multi} introduced FiA-Net, a fusion-with-attention model that integrates both intensity and RAVENS channels from sMRI data. The attention mechanism enabled the model to focus on the most informative regions, thereby improving its predictive performance. Later, He et al. \cite{he2022global} proposed a global-local transformer model, which combines global contextual information with fine-grained local features extracted from patches of sMRI data. This approach employs self-attention mechanisms to enhance the representation of brain aging patterns by dynamically adjusting attention to both global and local aspects of the data.}

\updatedText{Beyond CNNs and attention-based methods, graph neural networks (GNNs) have also been explored for brain age estimation, primarily due to their ability to model the brain as a network of regions of interest (ROIs). Pina et al. \cite{pina2022structural} proposed a GNN-based model that represents the brain as a graph, where ROIs are treated as nodes, and the dependencies between regions are modeled as edges. This structural representation allows for the prediction of brain age by capturing the interdependencies between different brain regions. Similarly, Xu et al. \cite{xu2021brain} used a graph-based model on diffusion tensor imaging (DTI) data to study the structural connectivity of brain aging, showing the potential of graph models in capturing complex relationships between brain regions.}

\updatedText{The integration of multimodal MRI data has emerged as a promising approach to enhance the accuracy of brain age estimation. Structural MRI (sMRI) and functional MRI (fMRI) provide complementary information, where sMRI captures structural features, and fMRI offers insights into functional connectivity. Early studies, such as those by Irimia et al. \cite{irimia2015statistical}, combined cortical thickness from sMRI with structural connectivity features to predict brain age using multivariate regression. Liem et al. \cite{liem2017predicting} proposed a stacked multimodal approach that integrates cortical anatomy from sMRI with functional connectivity derived from resting-state fMRI, demonstrating improved brain age prediction performance. Cherubini et al. \cite{cherubini2016importance} further investigated the fusion of T1-weighted MRI, T2-relaxometry, and fMRI using a voxel-based multiple regression model, highlighting the potential of multimodal data in capturing brain aging patterns. }

\updatedText{Building upon these foundational approaches, Cole et al. \cite{cole2020multimodality} developed a more comprehensive multimodal brain age model by integrating various MRI modalities, including T1-weighted MRI, T2-FLAIR, T2*, and diffusion MRI, along with resting-state fMRI. Although this approach demonstrated promising results, it relied on hand-crafted features and complex preprocessing steps, which limited its adaptability to new datasets. More recently, Mouches et al. \cite{mouches2022multimodal} proposed a fusion model combining sMRI with time-of-flight magnetic resonance angiography (TOF MRA) data using a simple fully convolutional network (SFCN) alongside linear regression (LR), thus simplifying the integration process while maintaining robust performance.}

Though recent advancements, traditional multimodal approaches often rely on simply concatenating features from different modalities without explicitly modeling their interactions. This limitation can overlook the potential synergies between sMRI and fMRI data that could enhance prediction accuracy. To address this, He et al. \cite{he2022global} proposed a global-local transformer model that combines global and local features through an attention mechanism, significantly improving performance. Similarly, Armanious et al. \cite{armanious2021age} incorporated both chronological and biological age information into CNN-based models, enhancing brain age prediction. Liu et al. \cite{liu2023risk} further highlighted the importance of demographic factors, such as gender, in brain age prediction, using a support vector regression (SVR) approach. Dular et al. \cite{dular2024base} extended brain age estimation to multisite data, achieving an MAE of 3.25 ± 2.70 years, demonstrating the value of large-scale, heterogeneous datasets in improving prediction accuracy. Additionally, the concept of disentangled representation learning has gained attention, with Cai et al. \cite{cai2023graph} employing a two-stream convolutional autoencoder to disentangle modality-specific features, improving upon traditional autoencoder designs. A concise summary of the existing literature is provided in Table \ref{litrature_tab}, highlighting key methodologies such as modality integration, sex-aware modeling, adversarial learning, and disentangled autoencoders. Studies utilizing attention-based and transformer networks for improved segmentation, particularly in challenging modalities, have shown promise in advancing model-based approaches \cite{farooq2023residual, kanwal2023mask, usman2024intelligent,usman2023mesaha, usman2024meds, rehman2023selective, iqbal2023ldmres, usman2023deha, usman2022dual, ullah2023ssmd, latif2018automating, lee2021evaluation, latif2018mobile, ullah2023mtss, ullah2023densely, usman2017using, ullah2022cascade, usman2020retrospective, usman2020volumetric, latif2018cross, latif2018phonocardiographic, latif2020leveraging,farooq2024lssf,naveed2024ad,iqbal2025tbconvl,iqbal2024tesl, usman2019motion,usman2022meds} \cite{naveed2024ra}. Autoencoders (AE) have been widely explored for multimodal fusion, with early and late fusion \cite{bib_20, farooq2023dc}. However, traditional AEs often struggle to differentiate between shared and complementary information, and noisy modalities can negatively impact latent representation learning across modalities.

\begin{figure*}[t]
\centering
\includegraphics[width=1\textwidth]{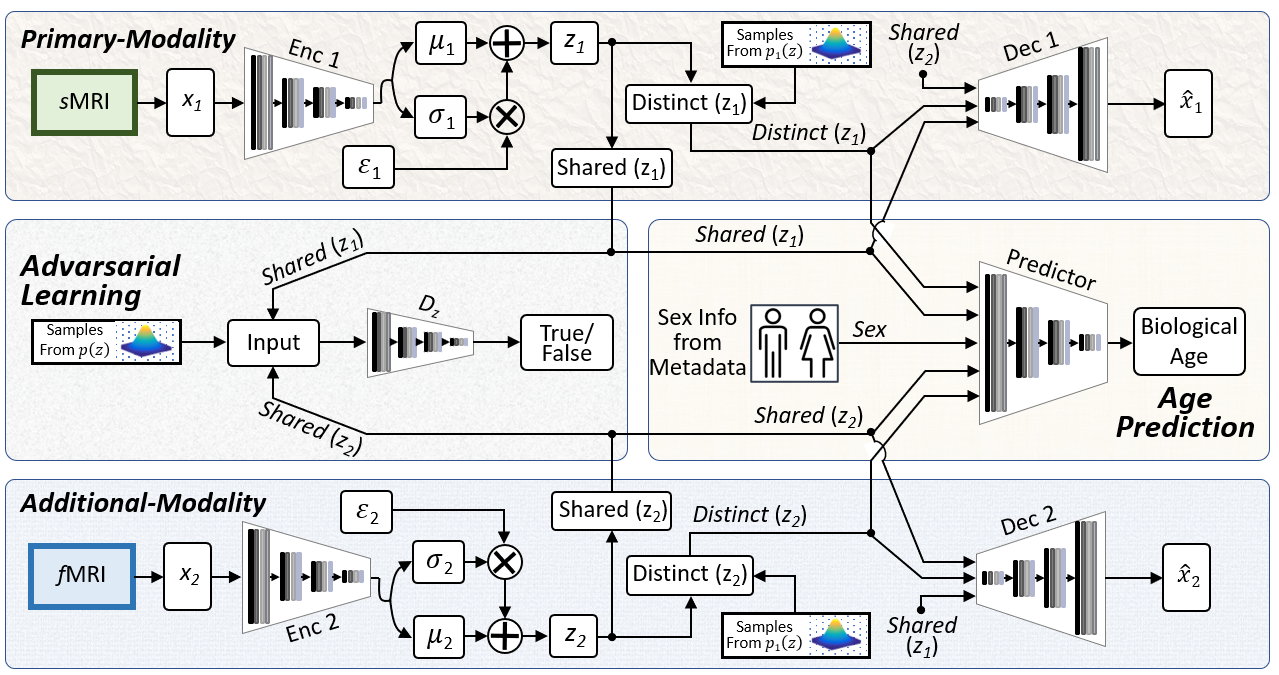}
\caption{\updatedText{Architecture of the proposed Multimodal Sex-Aware Adversarial Variational Autoencoder (SA-AVAE) for predicting biological brain age, utilizing sMRI as a compulsory modality and fMRI as an optional input to enhance prediction performance.}}
\label{proposed_architecture}
\end{figure*}

Despite recent advancements, current approaches to brain age estimation still face significant challenges, particularly in fully exploiting the interactions between multimodal data and accounting for sex-specific aging patterns \cite{usman2024advancing}. To address these issues, the proposed framework in this paper integrates sMRI and fMRI data while incorporating sex-specific information into the latent space, capturing both modality-specific and shared features through disentangled representations. Our framework, the Sex-Aware Adversarial Variational Autoencoder (SA-AVAE), not only improves predictive accuracy but also enhances model interpretability by explicitly considering demographic factors such as sex. The key contributions of our work include: 1) the introduction of a novel framework that integrates sMRI and fMRI data with sex information to improve accuracy and robustness; 2) the use of disentangled latent representations, achieved by applying adversarial and variational autoencoder losses to ensure effective separation of modality-specific and shared features; 3) a new loss weighting strategy for fine-tuning model parameters, providing insights into optimizing architectures for broader applications; and 4) extensive evaluation showing that our framework outperforms state-of-the-art methods, establishing a new benchmark for brain age estimation. This work paves the way for more effective, interpretable models in neuroimaging and brain health assessment, emphasizing the importance of integrating multimodal data, disentangled representations, and demographic factors to improve both prediction accuracy and clinical applicability.

\section{Proposed Methodology}
\label{pro}

We propose a multimodal framework, the Sex-Aware Adversarial Variational Autoencoder (SA-AVAE), which integrates adversarial and variational learning to disentangle features derived from sMRI and fMRI scans. These features are further combined with sex information to estimate biological brain age. While sMRI is more readily available, the framework is designed to function seamlessly even without the incorporation of fMRI, relying solely on sMRI when necessary. The components of the proposed framework are described in detail in the following subsections.

\subsection{\updatedText{Overall Framework}}

\updatedText{The architecture of the SA-AVAE framework is depicted in Figure \ref{proposed_architecture}. It consists of two autoencoder networks: the primary-modality path and the additional-modality path. Both networks share an identical architecture and are designed to process and integrate features extracted from sMRI (the primary modality) and fMRI (the optional modality).}

\updatedText{For each modality, a multi-layer perceptron (MLP) neural network serves as the encoder $E_i$, where $i = 1$ corresponds to sMRI and $i = 2$ corresponds to fMRI. The encoder generates a latent representation $z_i$ from the input feature vector $x_i$:}
\begin{equation}
z_i = E_i(x_i), \quad i \in \{1, 2\}.
\end{equation}
\updatedText{The latent representation $z_i$ is disentangled into two components:}
\begin{equation}
z_i = \left[\text{Shared}(z_i), \text{Dist}(z_i)\right],
\end{equation}
\updatedText{where $\text{Shared}(z_i)$ captures the common, modality-invariant information, and $\text{Dist}(z_i)$ encodes modality-specific features. The disentanglement is guided by the following principles:}
\begin{itemize}
    \item \updatedText{The concatenation of $\text{Shared}(z_i)$ and $\text{Dist}(z_i)$ reconstructs the original latent vector $z_i$:}
    \begin{equation}
    z_i = \text{Concat}\left(\text{Shared}(z_i), \text{Dist}(z_i)\right).
    \end{equation}
    \item \updatedText{Shared representations $\text{Shared}(z_1)$ and $\text{Shared}(z_2)$ should be as similar as possible to maximize commonality:}
    \begin{equation}
    \left\|\text{Shared}(z_1) - \text{Shared}(z_2)\right\|_2 \to 0.
    \end{equation}
    \item \updatedText{Distinct representations $\text{Dist}(z_1)$ and $\text{Dist}(z_2)$ should be as dissimilar as possible to emphasize modality-specific information:}
    \begin{equation}
    \left\|\text{Dist}(z_1) - \text{Dist}(z_2)\right\|_2 \to \max.
    \end{equation}
\end{itemize}

\updatedText{The disentangled latent features are fed into modality-specific decoders $Dec_i$ to reconstruct the input features. Finally, the concatenated latent spaces from both autoencoders, along with sex information, are passed into a regressor network $\mathbf{P}$ to estimate biological brain age:}
\begin{equation}
\mathcal{M} = \left[\text{Shared}(z_1), \text{Shared}(z_2), \text{Dist}(z_1), \text{Dist}(z_2), \text{Sex}\right].
\end{equation}

\subsection{\updatedText{Feature Disentanglement Strategy}}

\updatedText{The proposed SA-AVAE framework introduces a robust feature disentanglement strategy, leveraging adversarial learning, variational constraints, shared-distinct distance ratio loss, and cross-modality reconstruction to effectively separate shared and modality-specific latent representations. This comprehensive strategy ensures that the model captures complementary and modality-invariant features while maintaining the unique characteristics of each modality, thus optimizing its performance for multimodal brain age estimation.}

\subsubsection{\updatedText{Adversarial Learning for Shared Representations}}

\updatedText{Adversarial learning plays a pivotal role in aligning the shared latent representations $\text{Shared}(z_1)$ and $\text{Shared}(z_2)$ with a predefined prior distribution $p(\mathbf{z})$. This alignment ensures that the shared features are modality-invariant and contain common information across sMRI and fMRI data. A discriminator network $D_z$ is trained to distinguish between samples from the aggregated posterior distribution $q(\mathbf{z})$ and the prior $p(\mathbf{z})$. The adversarial loss for each modality is defined as:}
\begin{equation}
\updatedText{\mathcal{L}_{\text{adv}}^i = \mathbb{E}_{\mathbf{x}_i \sim p_d(\mathbf{x}_i)} \log \left(1 - D_z\left(\text{Shared}(E_i(\mathbf{x}_i))\right)\right) + \mathbb{E}_{\mathbf{z}_i \sim p(\mathbf{z})} \log \left(D_z(\mathbf{z}_i)\right),}
\end{equation}
\updatedText{where $i \in \{1, 2\}$ represents the modality index. The total adversarial loss is the sum of the losses across both modalities:}
\begin{equation}
\updatedText{\mathcal{L}_{\text{adv}} = \mathcal{L}_{\text{adv}}^1 + \mathcal{L}_{\text{adv}}^2.}
\end{equation}

\updatedText{This mechanism ensures that the shared representations are well-regularized and contribute to the effective disentanglement of modality-specific and shared features.}

\subsubsection{\updatedText{Variational Learning for Distinct Representations}}

\updatedText{To regularize the modality-specific latent spaces, variational learning enforces the distinct representations $\text{Dist}(z_1)$ and $\text{Dist}(z_2)$ to conform to predefined modality-specific prior distributions $p_1(\mathbf{z})$ and $p_2(\mathbf{z})$. This constraint ensures that the distinct latent spaces capture unique, modality-specific information. The variational loss for each modality is expressed as the Kullback-Leibler (KL) divergence:}
\begin{equation}
\updatedText{\mathcal{L}_{\text{var}}^i = \mathcal{D}_{\text{KL}}\left(q\left(\text{Dist}(z_i) | \mathbf{x}_i\right) \| p_i\left(\text{Dist}(z_i)\right)\right),}
\end{equation}
\updatedText{where $p_i\left(\text{Dist}(z_i)\right)$ is typically modeled as a Gaussian distribution $\mathcal{N}(\mathbf{0}, \mathbf{I})$. The total variational loss is defined as:}
\begin{equation}
\updatedText{\mathcal{L}_{\text{var}} = \mathcal{L}_{\text{var}}^1 + \mathcal{L}_{\text{var}}^2.}
\end{equation}

\updatedText{This variational constraint encourages the distinct representations to remain disentangled from the shared latent space, thereby improving the clarity of modality-specific features.}

\subsubsection{\updatedText{Shared-Distinct Distance Ratio Loss}}

\updatedText{To reinforce the separation between shared and distinct representations, the model employs a shared-distinct distance ratio loss, $\mathcal{L}_{\mathcal{D}}$. This loss emphasizes the disentanglement by balancing the similarity of shared representations and the dissimilarity of distinct representations. The loss is defined as:}
\begin{equation}
\mathcal{L}_{\mathcal{D}} = \frac{\mathcal{L}_{\mathcal{D}}^{\text{Shared}}}{\mathcal{L}_{\mathcal{D}}^{\text{Dist}}},
\end{equation}
where,
\begin{equation}
\mathcal{L}_{\mathcal{D}}^{\text{Shared}} = \mathbb{E}_{\mathbf{x}_1, \mathbf{x}_2}\left\|\text{Shared}\left(\mathbf{Enc}_1\left(\mathbf{x}_1\right)\right) - \text{Shared}\left(\mathbf{Enc}_2\left(\mathbf{x}_2\right)\right)\right\|_2,
\end{equation}
and
\begin{equation}
\mathcal{L}_{\mathcal{D}}^{\text{Dist}} = \mathbb{E}_{\mathbf{x}_1, \mathbf{x}_2}\left\|\text{Dist}\left(\mathbf{Enc}_1\left(\mathbf{x}_1\right)\right) - \text{Dist}\left(\mathbf{Enc}_2\left(\mathbf{x}_2\right)\right)\right\|_2.
\end{equation}

\updatedText{This ratio loss ensures a clear separation between shared and modality-specific features, enabling the model to disentangle complex multimodal information effectively.}

\subsubsection{\updatedText{Cross-Modality Reconstruction}}

\updatedText{To further encourage disentanglement, the framework incorporates a cross-modality reconstruction mechanism. This criterion leverages the shared representation from one modality to reconstruct the input of the other modality, ensuring that the shared features truly represent common information. Specifically:}
\begin{equation}
\updatedText{x_i \approx Dec_i\left(\text{Shared}(z_j), \text{Dist}(z_i)\right), \quad i \neq j.}
\end{equation}

\updatedText{The reconstruction loss for modality $i$ is defined as:}
\begin{equation}
\updatedText{\mathcal{L}_{\text{rec}}^i = \mathbb{E}_{\mathbf{x}_i \sim p_d(\mathbf{x}_i)} \left\|\mathbf{x}_i - Dec_i\left(\text{Shared}(E_j(\mathbf{x}_j)), \text{Dist}(E_i(\mathbf{x}_i))\right)\right\|_2^2.}
\end{equation}

\updatedText{The total reconstruction loss is the sum of the reconstruction losses across both modalities:}
\begin{equation}
\updatedText{\mathcal{L}_{\text{rec}} = \mathcal{L}_{\text{rec}}^1 + \mathcal{L}_{\text{rec}}^2.}
\end{equation}

\subsubsection{\updatedText{Comprehensive Objective Function}}
\updatedText{The complete objective function integrates all the discussed losses, balancing them with empirically determined trade-off parameters:}
\begin{equation}
\updatedText{\mathcal{L}_{\text{total}} = \mu_1 \mathcal{L}_{\text{adv}} + \mu_2 \mathcal{L}_{\text{var}} + \mu_3 \mathcal{L}_{\text{rec}} + \mu_4 \mathcal{L}_{\text{reg}} + \mu_5 \mathcal{L}_{\mathcal{D}},}
\end{equation}
\updatedText{where $\mu_1, \mu_2, \mu_3, \mu_4$, and $\mu_5$ are weighting coefficients. The regression loss $\mathcal{L}_{\text{reg}}$ measures the discrepancy between the predicted and actual brain age using the L2 norm:}
\begin{equation}
\updatedText{\mathcal{L}_{\text{reg}} = \mathbb{E}_{\mathbf{x}_1, \mathbf{x}_2}\left\|y - \mathbf{P}\left(M\left(\mathbf{x}_1, \mathbf{x}_2\right)\right)\right\|_2.}
\end{equation}

\updatedText{This comprehensive approach ensures robust disentanglement, accurate reconstruction, and precise brain age estimation, making the SA-AVAE framework highly effective for multimodal neuroimaging analysis.}

\subsection{\updatedText{Single Modality Algorithm}}
\label{unimodal_algo}
\updatedText{In practical scenarios, the availability of datasets containing both sMRI and fMRI modalities is often limited, with most samples including only sMRI data. To address this imbalance, the SA-AVAE framework is adapted to function effectively in a single-modality setting, particularly for sMRI. This adaptation involves deactivating the fMRI encoder-decoder path and modifying the overall objective function to optimize performance with only sMRI input.}

\subsubsection{\updatedText{Adaptation of the Objective Function}}

\updatedText{To accommodate the absence of the fMRI modality, the objective function is redefined. Adversarial and variational losses are applied exclusively to the sMRI path, while the regression and reconstruction losses are adjusted to rely solely on sMRI-derived representations. The revised losses are as follows:}

\begin{itemize}
    \item \updatedText{\textbf{Adversarial Loss:} Ensures alignment of shared latent representations with a prior distribution:}
    \[
    \updatedText{\mathcal{L}^{'}_{\text{adv}} = \mathcal{L}_{\text{adv}}^1.}
    \]
    \item \updatedText{\textbf{Variational Loss:} Regularizes the distinct latent space to conform to a predefined prior:}
    \[
    \updatedText{\mathcal{L}^{'}_{\text{var}} = \mathcal{L}_{\text{var}}^1.}
    \]
    \item \updatedText{\textbf{Regression Loss:} Predicts biological brain age using sMRI-derived latent representations:}
    \[
    \updatedText{\mathcal{L}^{'}_{\text{Reg}} = \mathbb{E}_{\mathbf{x}_1}\left\|y - \mathbf{P}\left(\mathcal{M}\left(\mathbf{x}_1\right)\right)\right\|_2^2,}
    \]
    \updatedText{where $\mathcal{M}(\mathbf{x}_1)$ includes the shared and distinct representations along with sex information.}
    \item \updatedText{\textbf{Reconstruction Loss:} Ensures accurate reconstruction of the sMRI input:}
    \[
    \updatedText{\mathcal{L}^{'}_{\text{Rec}} = \mathbb{E}_{\mathbf{x}_1}\left\|\mathbf{x}_1 - \mathbf{Dec}_1\left(\text{Shared}\left(\mathbf{Enc}_1\left(\mathbf{x}_1\right)\right), \text{Dist}\left(\mathbf{Enc}_1\left(\mathbf{x}_1\right)\right)\right)\right\|_2^2.}
    \]
\end{itemize}

\updatedText{The modified objective function is:}
\begin{equation}
\updatedText{\mathcal{L}^{'}_{\mathbf{Enc}_1, \mathbf{Dec}_1, \mathbf{P}} = \eta_1 \mathcal{L}^{'}_{\text{Reg}} + \eta_2 \mathcal{L}^{'}_{\text{Rec}} + \eta_3 \mathcal{L}^{'}_{\text{adv}} + \eta_4 \mathcal{L}^{'}_{\text{var}},}
\label{single_modality_objective}
\end{equation}
\updatedText{where $\eta_1, \eta_2, \eta_3$, and $\eta_4$ are trade-off parameters controlling the contribution of each loss term.}

\subsubsection{\updatedText{Training Strategy for Uni-modal Setting}}

\updatedText{The single-modality training procedure for the SA-AVAE framework has been adapted to ensure optimal performance using only sMRI data, addressing the challenges posed by limited multimodal datasets. During training, the encoder $Enc_1$ processes the sMRI input $\mathbf{x}_1$ to generate latent representations that are disentangled into shared and distinct components. These representations are defined as $\text{Shared}(z_1)$ and $\text{Dist}(z_1)$, respectively, and together they form the latent space $z_1 = \left[\text{Shared}(z_1), \text{Dist}(z_1)\right]$. The decoder $Dec_1$ then reconstructs the original input $\mathbf{x}_1$ by leveraging these latent features, ensuring that the reconstructed data closely matches the input.}

\updatedText{To enhance the learned representations, adversarial regularization is applied to the shared latent space $\text{Shared}(z_1)$, aligning it with a predefined prior distribution via a discriminator network. Simultaneously, variational regularization is employed on the distinct latent space $\text{Dist}(z_1)$ to encourage alignment with a modality-specific prior distribution using the KL divergence. The regressor network $\mathbf{P}$ predicts the biological brain age $y$ by combining the shared and distinct features with additional sex information encoded in $\mathcal{M}(\mathbf{x}_1)$. The parameters of the encoder, decoder, regressor, and discriminator are updated jointly using the modified objective function, which balances regression, reconstruction, adversarial, and variational losses.}

\updatedText{This approach not only ensures robust feature learning and disentanglement but also provides several benefits. First, the framework is highly adaptable, allowing it to operate seamlessly with datasets containing only sMRI data, which is crucial for real-world applications where multimodal data is often unavailable. Second, the absence of the fMRI path reduces computational overhead, resulting in faster training and inference, which enhances efficiency. Third, adversarial and variational regularizations ensure that the learned representations remain robust and meaningful, even in the single-modality scenario. Finally, the modular design of the framework ensures scalability, allowing future integration with additional modalities or extension to other single-modality tasks. These attributes make the single-modality adaptation of the SA-AVAE framework a powerful and flexible solution for brain age estimation in diverse settings.}

\begin{figure*}[ht]
\centering
\includegraphics[width=1\textwidth]{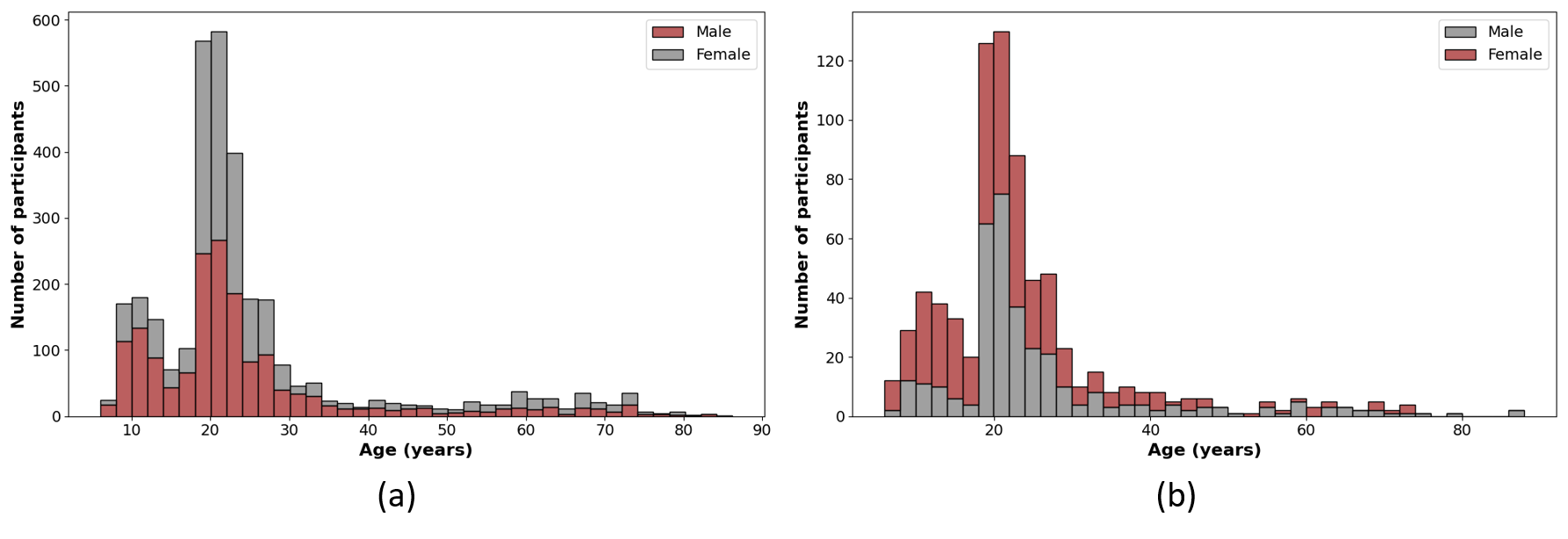}
\caption{\updatedText{Age distribution of male and female participants in the training and validation sets of the OpenBHB dataset\cite{dufumier2022openbhb} , shown in (a) and (b), respectively.}}
\label{data_distribution}
\end{figure*}

\section{Experimental Setup}
\label{es_section}
\subsection{Dataset and Pre-processing}

In our experiments, we utilized the OpenBHB dataset \cite{dufumier2022openbhb}, a large-scale and comprehensive resource comprising 5330 3D brain MRI scans collected from 71 different sites. This diverse dataset includes scans from a wide range of demographics, ensuring broad genetic and geographical representation. Specifically, 3984 of the scans are publicly available, with 3227 scans allocated for training and 757 scans designated for validation. The validation set is further split into two subsets: 362 internal test samples and 390 external test samples, allowing for a comprehensive evaluation of the model's performance on both internal and unseen data. 

The OpenBHB dataset is organized into 10 distinct sub-datasets, which include subjects from varied ethnic backgrounds, including European-American, European, and Asian descent. This diversity ensures that our model is trained and evaluated on a broad spectrum of genetic backgrounds, enhancing its robustness and generalizability across different populations. The age range of the subjects spans from 16 to 86 years, and the dataset exhibits a balanced distribution of male and female subjects across different age groups, as shown in Figure \ref{data_distribution}. This balanced sex distribution is crucial for minimizing biases and ensuring that the trained model can generalize effectively across both male and female populations.

For the single-modality experiments with the SA-AVAE framework, we utilized the full publicly available OpenBHB dataset, which consists predominantly of structural MRI (sMRI) data. This extensive dataset enabled us to train the model effectively in single-modality settings, focusing exclusively on sMRI data. For multimodal experiments, we selected subsets of the dataset that include both sMRI and functional MRI (fMRI) data. Specifically, we incorporated two smaller datasets, referenced in \cite{dataset1} and \cite{dataset2}, which contain 66 and 315 scans, respectively. These two subsets were combined to create a unified multimodal dataset, resulting in a total of 381 scans. This approach allowed us to explore the impact of multimodal data integration on brain age estimation and assess the performance of our framework in leveraging both structural and functional brain information.

Overall, the diversity of the OpenBHB dataset, combined with its large size and rich multimodal data, provides a solid foundation for evaluating our proposed framework. The varied demographic representation and the inclusion of both single-modality and multimodal data ensure that our experiments are not only comprehensive but also robust across different populations and imaging modalities.

The high dimensionality of MRI scans, combined with the limited size of available datasets, poses significant computational challenges and increases the risk of overfitting when directly used in neural network training. To address these challenges, an initial feature selection step is essential. Feature selection methods are broadly categorized into three types based on their interaction with the learning model: filter methods, wrapper methods, and embedded methods \cite{urbanowicz2018relief}. We adopted a filter method, due to its independence from specific models and computational efficiency, and its proven effectiveness\cite{hu2020disentangled}.

For feature selection, we employed the Random Forest algorithm, which is well-suited for handling high-dimensional data with highly correlated features \cite{kawakubo2012rapid}. This process results in $m_1$ and $m_2$ selected features from the first (sMRI) and second (fMRI) modalities, respectively. The dataset after feature selection can be mathematically represented as:
\begin{equation}
    (X_1, X_2, Y) = \{(X_{11}, X_{21}, y_1), \ldots, (X_{1n}, X_{2n}, y_n)\},
\end{equation}
where:
\begin{itemize}
    \item $X_{1n} \in \mathbb{R}^{m_1}$ represents the feature vector of the $n$-th instance derived from the sMRI modality.
    \item $X_{2n} \in \mathbb{R}^{m_2}$ represents the feature vector of the $n$-th instance derived from the fMRI modality.
    \item $y_n \in \mathbb{R}$ corresponds to the target outputs (e.g., age and gender) for the $n$-th instance.
    \item $N$ denotes the total number of instances in the dataset.
\end{itemize}

The resulting feature vectors $X_1$ and $X_2$, along with the target outputs $Y$, form the basis for training and evaluating the proposed model.

\subsection{Model Architecture and Training Strategy}

\begin{figure*}[ht]
\centering
\includegraphics[width=0.9\textwidth]{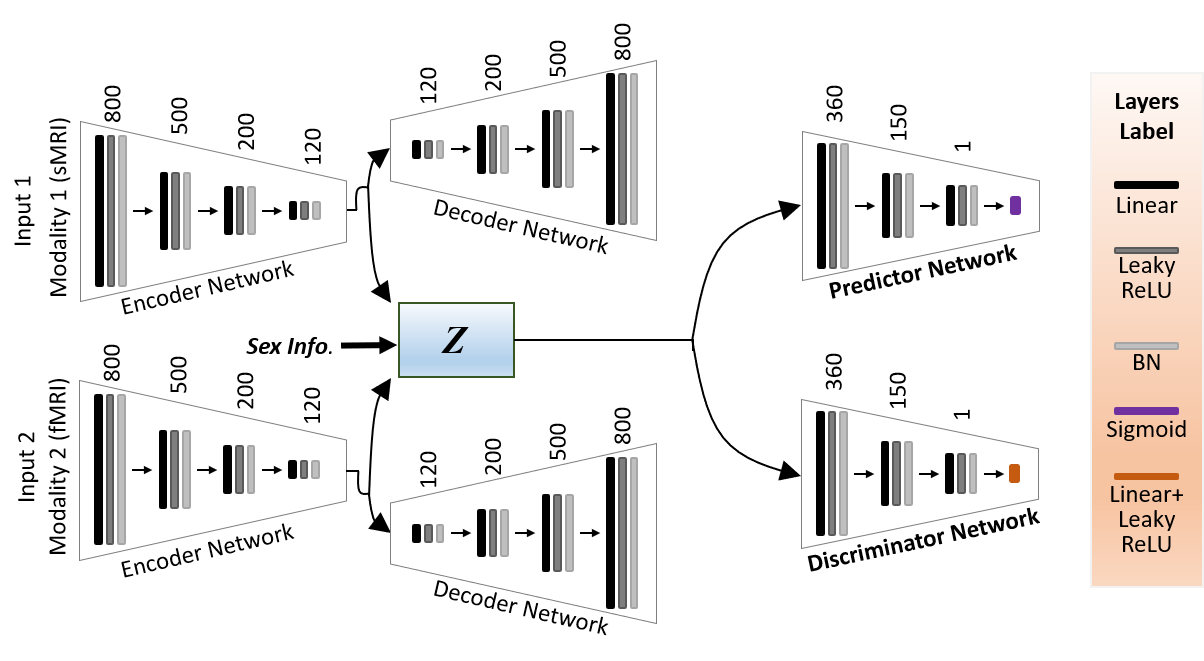}
\caption{\updatedText{Illustration of architectural details of our proposed Sex-Aware Adversarial Variational Autoencoder.}}
\label{backbone_net}
\end{figure*}

The architecture of the proposed Sex-Aware Adversarial Variational Autoencoder (SA-AVAE), employed in our experiments, is depicted in Figure \ref{backbone_net}. This architecture integrates encoders and decoders for both sMRI and fMRI modalities, alongside a shared regressor and a discriminator to facilitate adversarial learning. To ensure reproducibility and optimize the model's performance, several key hyperparameters were carefully selected. First, the batch size was set to 20 during training to balance memory usage and model convergence. The latent space of the model was designed with a total dimensionality of 120, divided into a shared space of 50 dimensions and modality-specific spaces of 70 dimensions each. These dimensions were determined empirically, striking a balance between the representational capacity of the model and the computational efficiency required for large-scale experiments. For training, the Adam optimizer was used with an initial learning rate of 0.001, which was dynamically adjusted; if no improvement in validation performance was observed after nine epochs, the learning rate was reduced to a quarter of its current value to aid in convergence. Additionally, an early stopping mechanism was implemented to mitigate overfitting, halting the training process when no improvement was seen in the validation set after a predefined number of epochs. The training and evaluation of the SA-AVAE framework were conducted on an NVIDIA RTX 4090 GPU using the Keras library with a TensorFlow backend. The model had a total of 2.7 million trainable parameters. Training took approximately 12 hours to complete, while testing was conducted in about 1.5 hours, demonstrating the model’s computational efficiency, which makes it suitable for handling large-scale neuroimaging datasets.

Several implementation strategies were critical for optimizing the performance of the SA-AVAE framework. During training, the reconstruction losses for both sMRI and fMRI modalities were balanced to ensure that each modality contributed equally to the shared latent representations, preventing bias toward one modality. The adversarial and variational losses were carefully scaled using empirically derived trade-off parameters. This scaling ensured that the shared and distinct latent spaces were effectively disentangled, allowing the model to learn complementary features from both modalities. Furthermore, the regressor was jointly trained with both the encoder and decoder networks to ensure that the latent representations were tailored for accurate age prediction. Dropout and weight decay were employed to eliminate the possibility of overfitting and improve generalization ability. These measures, in combination with robust preprocessing and careful feature selection, enabled the SA-AVAE framework to achieve competitive performance in biological brain age prediction, effectively leveraging multimodal neuroimaging data.

\begin{table*}[ht]
\centering
\caption{\updatedText{Performance comparison of different models evaluated using Mean Absolute Error (\textit{MAE}), Root Mean Square Error (\textit{RMSE}), and Coefficient of Determination (\(R^2\)) for overall, male, and female scans. \textit{The best results are underlined.}}}
\begin{tabular}{@{}lccccccccc@{}}
\toprule
\updatedText{\textbf{Model Name}} & \multicolumn{3}{c}{\updatedText{\textbf{Overall}}} & \multicolumn{3}{c}{\updatedText{\textbf{Male}}} & \multicolumn{3}{c}{\updatedText{\textbf{Female}}} \\ 
\cmidrule(lr){2-4} \cmidrule(lr){5-7} \cmidrule(lr){8-10}
 & \updatedText{\textit{MAE} $\downarrow$} & \updatedText{\textit{RMSE} $\downarrow$} & \updatedText{\(R^2\uparrow\)} & \updatedText{\textit{MAE} $\downarrow$} & \updatedText{\textit{RMSE} $\downarrow$} & \updatedText{\(R^2\uparrow\)} & \updatedText{\textit{MAE} $\downarrow$} & \updatedText{\textit{RMSE} $\downarrow$} & \updatedText{\(R^2\uparrow\)} \\ 
\midrule
\updatedText{AE}            & \updatedText{3.590 $\pm$ 1.910} & \updatedText{4.067} & \updatedText{0.885} & \updatedText{3.695 $\pm$ 1.898} & \updatedText{4.154} & \updatedText{0.859} & \updatedText{3.394 $\pm$ 1.918} & \updatedText{3.899} & \updatedText{0.908} \\
\updatedText{AAE}           & \updatedText{3.279 $\pm$ 1.882} & \updatedText{3.781} & \updatedText{0.901} & \updatedText{3.419 $\pm$ 1.863} & \updatedText{3.894} & \updatedText{0.876} & \updatedText{3.017 $\pm$ 1.889} & \updatedText{3.560} & \updatedText{0.923} \\
\updatedText{VAE}           & \updatedText{3.142 $\pm$ 1.782} & \updatedText{3.612} & \updatedText{0.910} & \updatedText{3.212 $\pm$ 1.765} & \updatedText{3.665} & \updatedText{0.890} & \updatedText{3.011 $\pm$ 1.807} & \updatedText{3.512} & \updatedText{0.925} \\
\updatedText{AVAE}          & \updatedText{2.921 $\pm$ 1.608} & \updatedText{3.334} & \updatedText{0.923} & \updatedText{2.979 $\pm$ 1.594} & \updatedText{3.379} & \updatedText{0.907} & \updatedText{2.813 $\pm$ 1.626} & \updatedText{3.250} & \updatedText{0.936} \\
\updatedText{SA-AVAE}       & \updatedText{\underline{2.722 $\pm$ 1.351}} & \updatedText{\underline{3.039}} & \updatedText{\underline{0.936}} & \updatedText{\underline{2.727 $\pm$ 1.345}} & \updatedText{\underline{3.041}} & \updatedText{\underline{0.924}} & \updatedText{\underline{2.713 $\pm$ 1.363}} & \updatedText{\underline{3.036}} & \updatedText{\underline{0.944}} \\
\bottomrule
\end{tabular}
\label{tab_ablation}
\end{table*}

\section{Results and Discussion}
\label{res}
\subsection{\updatedText{Ablation Study}}
\updatedText{In this section, we perform an ablation study to evaluate the contribution of each component in the proposed framework. We implemented various downgraded versions of the multimodal model, including the Autoencoder (AE), Adversarial Auto-Encoder (AAE), Variational Auto-Encoder (VAE), Adversarial Variational Auto-Encoder (AVAE), and our proposed Sex-Aware Adversarial Variational Auto-Encoder (SA-AVAE). Table \ref{tab_ablation} summarizes the results of each model in terms of Mean Absolute Error (MAE), Root Mean Square Error (RMSE), and Coefficient of Determination (\(R^2\)) for overall, male, and female samples. The results indicate that the proposed SA-AVAE outperforms all downgraded variants. From Table \ref{tab_ablation}, it is evident that each added component contributes to the overall performance improvement achieved by SA-AVAE. Specifically, the inclusion of both adversarial and variational learning not only enhances the model's performance individually but also results in a substantial improvement when combined. Most importantly, the integration of sex information significantly boosts the performance of our proposed SA-AVAE framework.}

\subsection{\updatedText{Comparison with State-of-the-Art Methods}}

\updatedText{In this section, we compare the performance of our proposed framework with previously published methods. To ensure a fair comparison, we specifically shortlisted studies that utilized the same dataset, OpenBHB \cite{dufumier2022openbhb}, which is also used in this study. Table \ref{tab_SOTA_comparison} provides a summary of these studies along with their respective performances in biological brain age estimation, measured in terms of mean absolute error (MAE).}

\updatedText{Overall, our proposed method demonstrates superior performance compared to the previously published studies. Among the methods compared, Ahmed et al. \cite{ahmed2023robust} is the only one, apart from our framework, that utilized features extracted from 3D sMRI scans. Their approach aimed to improve brain age estimation by extracting region-wise features and classifying participants into distinct age groups or clusters. These region-wise features were later integrated into regression models. This method achieved the second-lowest MAE in the comparison. Aqil et al. \cite{aqil2023confounding} employed 3D T1-weighted MRI scans combined with healthy Montreal Neurological Institute (MNI) templates as inputs to a customized encoder-decoder architecture with an additional regressor branch. Despite their attempt to enforce the network to align with brain patterns by comparing them against templates, their method achieved the highest MAE in the comparison. This was likely due to the computational overhead introduced by the 3D network and its inability to fully capture the underlying brain patterns effectively.}

\updatedText{Similarly, Cheshmi et al. \cite{cheshmi2023brain} and Träuble et al. \cite{trauble2024contrastive} also utilized 3D inputs for brain age estimation. Cheshmi et al. \cite{cheshmi2023brain} employed a ResNet-18 model \cite{he2016deep} trained using federated learning, while Träuble et al. \cite{trauble2024contrastive} introduced a novel contrastive loss in their framework. Träuble et al.'s method dynamically adapted the contrastive loss during training to focus on localized neighborhoods of samples and incorporated brain stiffness—a mechanical property sensitive to age-related changes. This marked the first application of self-supervised learning to brain stiffness for brain age prediction. Despite their innovations, both methods lacked the integration of sex information, which negatively impacted their brain age estimation performance.}

\updatedText{In contrast, our framework surpasses these methods by leveraging a combination of adversarial learning and variational learning. It effectively integrates features extracted from both sMRI and fMRI modalities, while incorporating sex information to account for sex-specific aging patterns. The state-of-the-art performance of our framework can be attributed to its ability to disentangle the feature space, dividing the latent space into shared and distinct features. This enables the model to better capture modality-invariant and modality-specific information, leading to more accurate predictions. These advancements demonstrate the efficiency and robustness of our framework, making it suitable for real-time applications.}

\begin{table}[ht]
\centering
\caption{\updatedText{Comparison of brain age estimation performance across various methods, including the proposed SA-AVAE model. The table presents the Mean Absolute Error (MAE) for each method, along with a brief description of the algorithm/architecture used.}}
\begin{tabular}{lll}
\hline
\rowcolor[HTML]{EFEFEF} 
\multicolumn{1}{c}{\cellcolor[HTML]{EFEFEF}\textbf{Study, year}} &
  \multicolumn{1}{c}{\cellcolor[HTML]{EFEFEF}\textbf{Algorithm/Architecture Details}} &
  \multicolumn{1}{c}{\cellcolor[HTML]{EFEFEF}\textbf{MAE}} \\ \hline
Aqil \textit{et al.} \cite{aqil2023confounding}, 2023   & Modified Autoencoder & 4.554    \\ \hline
Ahmed \textit{et al.} \cite{ahmed2023robust}, 2023   & Classical Regression Algorithms & 3.250   \\ \hline
Cheshmi \textit{et al.} \cite{cheshmi2023brain}, 2023  & 3D ResNet-18 \cite{he2016deep} & 3.860   \\ \hline
Träuble \textit{et al.} \cite{trauble2024contrastive}, 2024 & 3D ResNet-18 + Contrastive Loss & 3.724 \\ \hline
Our Method       & SA-AVAE & 2.722 \\ \hline
\end{tabular}
\label{tab_SOTA_comparison}
\end{table}

\begin{figure*}[t]
    \centering
    \begin{subfigure}[t]{0.48\textwidth}
        \centering
        \includegraphics[width=\textwidth]{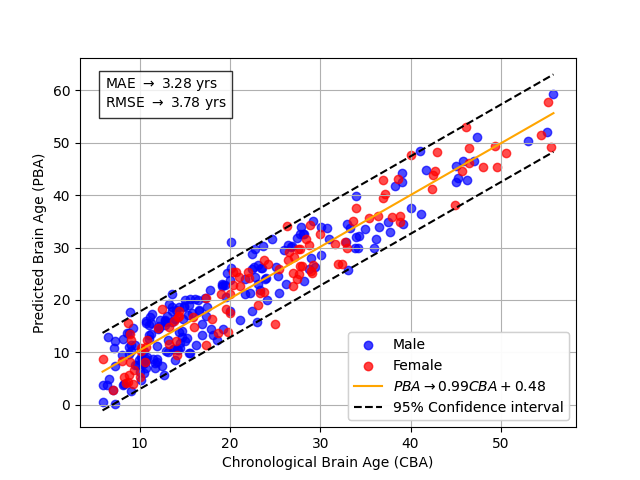}
        \caption{AAE}
    \end{subfigure}
    \begin{subfigure}[t]{0.48\textwidth}
        \centering
        \includegraphics[width=\textwidth]{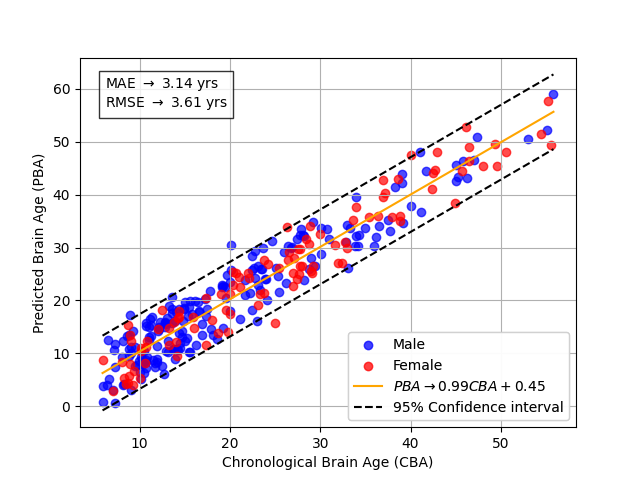}
        \caption{ VAE}
    \end{subfigure}
    \begin{subfigure}[t]{0.48\textwidth}
        \centering
        \includegraphics[width=\textwidth]{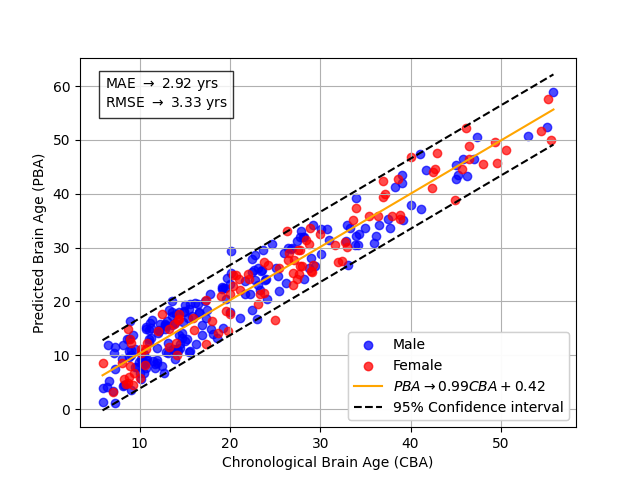}
        \caption{AVAE}
    \end{subfigure}
    \begin{subfigure}[t]{0.48\textwidth}
        \centering
        \includegraphics[width=\textwidth]{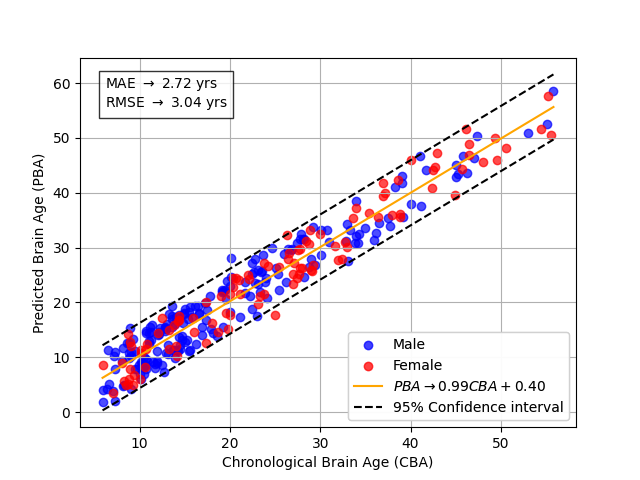}
        \caption{\textit{Our Proposed} SA-AVAE}
    \end{subfigure}
    
    \caption{\updatedText{Comparison of predictions for different multi-modal models: (a) AAE, (b) VAE, (c) AVAE, (d) SA-AVAE. Each graph plots predicted brain age versus chronological brain age, with gender and confidence intervals indicated.}}
    \label{models_comparison}
\end{figure*}

\subsection{Robustness Analysis}

\updatedText{To evaluate the robustness of the SA-AVAE model, we performed 10-fold cross-validation to assess the consistency of its performance across the dataset. Figure \ref{models_comparison} shows scatter plots comparing the predicted and chronological ages of the SA-AVAE and other models, under identical experimental conditions, highlighting the relative performance and robustness of each. The results demonstrate that combining adversarial and variational learning enhances both performance and robustness. Incorporating sex information further improves the model's robustness, as evidenced by lower standard deviation. Notably, the SA-AVAE model outperforms other variants in terms of confidence intervals, underscoring the importance of sex-aware inputs. Our model’s superior performance is attributed to its ability to create a disentangled latent space using three distinct prior distributions, enforced by adversarial and variational losses.}

To further illustrate the robustness of our proposed framework, Table \ref{results_breakdown_tab} presents a breakdown of the model's performance across various age groups. Among all models, the proposed SA-AVAE consistently outperforms others in every age group, highlighting the advantages of sex-aware disentangled learning. Ultimately, the SA-AVAE model delivers the most consistent and accurate predictions, reinforcing the efficacy of our approach.

\begin{table}[t]
\centering
\caption{\updatedText{Performance of various multimodal frameworks for biological brain age estimation across four age groups: G1 (under 25 years), G2 (25 to 35 years), G3 (35 to 45 years), and G4 (45 to 55 years), measured in terms of Mean Absolute Error (MAE).}}
\label{results_breakdown_tab}
\begin{tabular}{@{}lcccc@{}}
\toprule
    \textbf{Framework} & \textbf{G1} & \textbf{G2} & \textbf{G3} & \textbf{G4} \\
    \midrule
AE         & 3.63 ± 1.87 & 3.51 ± 1.96 & 4.01 ± 1.96 & 2.58 ± 1.79 \\
AAE        & 3.32 ± 1.86 & 3.20 ± 1.90 & 3.65 ± 1.97 & 2.31 ± 1.67 \\
VAE        & 3.17 ± 1.76 & 3.10 ± 1.80 & 3.53 ± 1.85 & 2.23 ± 1.64 \\
AVAE       & 2.94 ± 1.58 & 2.87 ± 1.63 & 3.28 ± 1.66 & 2.09 ± 1.50 \\
SA-AVAE    & \textbf{2.74 ± 1.32} & \textbf{2.68 ± 1.41} & \textbf{3.06 ± 1.36} & \textbf{2.04 ± 1.30} \\ \bottomrule
\end{tabular}
\end{table}

\subsection{\updatedText{Multi-Modality Analysis}}
\label{multimodal_analysis_section}
\begin{figure*}[t]
\centering
\includegraphics[width=1\textwidth]{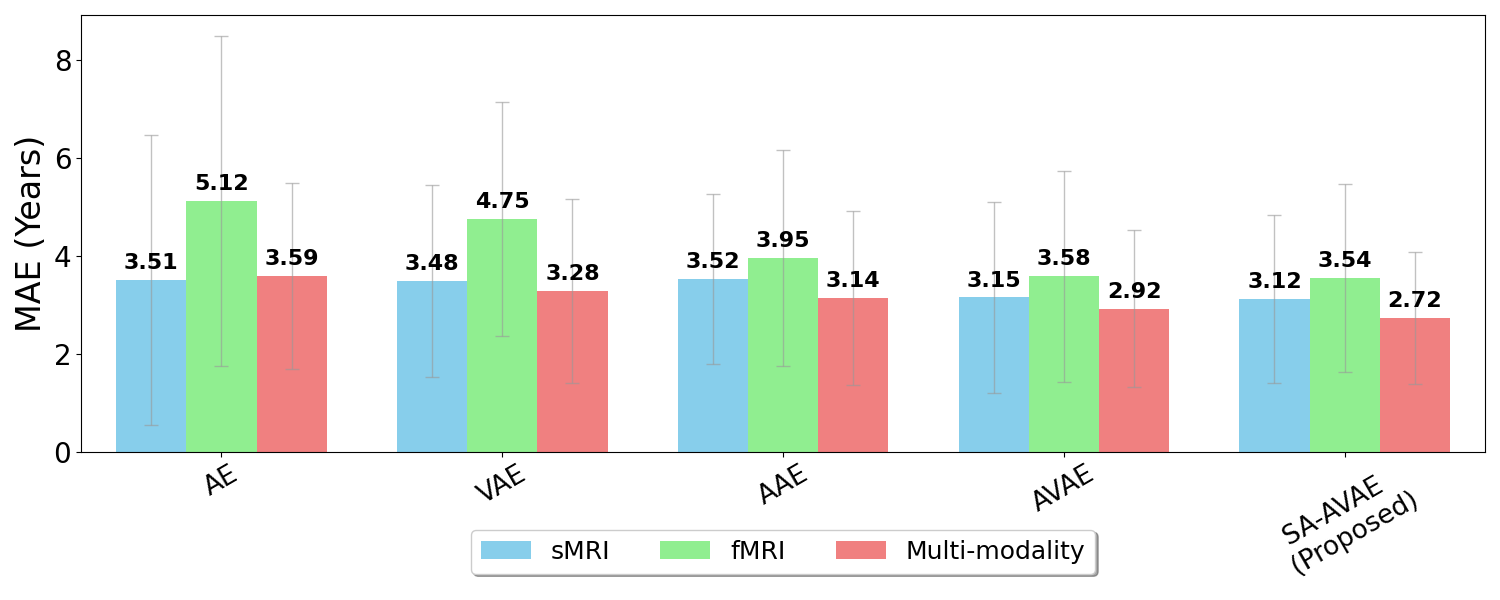}
\caption{\updatedText{Mean Absolute Error (MAE) for brain age estimation obtained from unimodal regressors using sMRI, fMRI, and a multimodal regressor combining both sMRI and fMRI.}}
\label{results_bar_chart}
\end{figure*}

\updatedText{The impact of multimodal fusion on model performance was comprehensively assessed by comparing unimodal (sMRI or fMRI alone) and multimodal (combined sMRI and fMRI) data strategies, as shown in Figure \ref{results_bar_chart}. This evaluation clearly demonstrated that unimodal models trained exclusively on fMRI data produced the least favorable results, indicating that fMRI alone does not provide sufficient discriminative power for accurate age prediction. In contrast, models trained on sMRI data, while still unimodal, exhibited moderately improved performance, suggesting that structural MRI alone holds more valuable information for estimating brain age.}

\updatedText{More notably, the integration of both sMRI and fMRI data into a multimodal framework resulted in a significant performance boost across all metrics, highlighting the clear advantages of combining complementary information from different modalities. Specifically, the MAE for age prediction using only sMRI data ranged from 3.52 to 2.72 years, showing reasonable accuracy. However, when relying solely on fMRI, the MAE ranged between 5.12 and 3.54 years, underperforming compared to sMRI. By fusing sMRI and fMRI, the MAE improved, yielding values between 3.59 and 2.72 years, demonstrating the enhanced predictive power that arises from the synergistic combination of structural and functional brain imaging data. These results further reinforce the critical role of multimodal data fusion in enhancing model accuracy and reliability in brain age estimation. It is important to note that the simple autoencoder network in the unimodal setting (i.e., using only sMRI as input) performed better than its multimodal counterpart. This can be attributed to the limited feature learning capability of traditional autoencoders. In contrast, all multimodal models, including the proposed SA-AVAE, showed consistent improvement over their unimodal counterparts, demonstrating the value of combining sMRI and fMRI data. The results indicate that although conventional multimodal fusion techniques can sometimes amplify noise, particularly from fMRI data, leading to a drop in accuracy compared to using sMRI alone, the SA-AVAE framework adeptly overcomes this limitation. By integrating both sMRI and fMRI data, the SA-AVAE model significantly enhanced performance, achieving a reduction in the MAE from 3.12 years with sMRI alone to 2.72 years with the multimodal approach. This improvement demonstrates the model’s superior ability to harmoniously combine different data modalities, thereby enhancing predictive precision without the usual drawback of noise interference.}

\updatedText{Additionally, we trained the proposed model in a unimodal setting (as described in Section \ref{unimodal_algo}) using all available sMRI scans from the OpenBHB dataset \cite{dufumier2022openbhb} to compare the performance with our multimodal SA-AVAE framework. Table \ref{modality_tab} summarizes the results from both models. Notably, the unimodal model was trained on 3,200 sMRI scans, whereas the multimodal framework used features extracted from just 320 sMRI and fMRI scans. Although the unimodal model was trained on ten times more data, which helped it to improve its performanc for brain age estimation (i.e., by improving MAE 3.12 to 2.91), the multimodal SA-AVAE still outperformed its unimodal counterpart. This can be attributed to the additional value provided by the fMRI features, which enhance the model’s ability to learn meaningful information crucial for understanding brain aging patterns. Thus, we conclude that the proposed multimodal framework can achieve comparable performance even with a smaller dataset.}

\begin{table}[]
\caption{\updatedText{Comparison of SA-AVAE performance with multimodal and unimodal data, evaluated on different metrics for overall, male, and female groups.}}
\begin{tabular}{|cc|c|c|}
\hline
\rowcolor[HTML]{EFEFEF} 
\multicolumn{2}{|c|}{\cellcolor[HTML]{EFEFEF}\textbf{Model Type}} &
  \textbf{\begin{tabular}[c]{@{}c@{}}Multimodal SA-AVAE \\ \end{tabular}} &
  \textbf{\begin{tabular}[c]{@{}c@{}}Unimodal SA-AVAE\\ \end{tabular}} \\ \hline
\multicolumn{1}{|c|}{}                                   & MAE  & 2.722 ± 1.351   & 2.906 ± 1.528   \\ \cline{2-4} 
\multicolumn{1}{|c|}{}                                   & RMSE & 3.039           & 3.283           \\ \cline{2-4} 
\multicolumn{1}{|c|}{\multirow{-3}{*}{\textbf{Overall}}} & R2   & 0.936           & 0.925           \\ \hline
\multicolumn{1}{|c|}{}                                   & MAE  & 2.727 ± 1.345   & 2.956 ± 1.519   \\ \cline{2-4} 
\multicolumn{1}{|c|}{}                                   & RMSE & 3.041           & 3.324           \\ \cline{2-4} 
\multicolumn{1}{|c|}{\multirow{-3}{*}{\textbf{Male}}}    & R2   & 0.924           & 0.91            \\ \hline
\multicolumn{1}{|c|}{}                                   & MAE  & 2.713 ± 1.363   & 2.812 ± 1.539   \\ \cline{2-4} 
\multicolumn{1}{|c|}{}                                   & RMSE & 3.036           & 3.206           \\ \cline{2-4} 
\multicolumn{1}{|c|}{\multirow{-3}{*}{\textbf{Female}}}  & R2   & 0.944           & 0.938           \\ \hline
\end{tabular}
\label{modality_tab}
\end{table}

\subsection{Impact of Incorporating Sex Information}

In this section, we analyze the impact of incorporating sex information into the brain age estimation model. Specifically, we compare the performance of the proposed adversarial variational autoencoder (AVAE) network with and without sex information. To expand our analysis, we explore two methods of incorporating sex information. The first method involves employing multitask learning, where the model is required to predict the subject's sex along with biological brain age estimation. The second method, which we adopt in the proposed SA-AVAE framework, directly inputs the sex information into the regressor.

Table \ref{sex_info_tab} summarizes the results obtained from three different frameworks for brain age estimation, evaluating male, female, and overall groups, along with the number of trainable parameters. It can be observed that models incorporating sex information outperform the model without sex information across all groups, highlighting the significance of including sex information in the model. Furthermore, the inclusion of sex information via multitask learning significantly improves the network's performance, although the improvement is slightly more pronounced in one gender group (female or male) than the other. On the other hand, the framework with direct inclusion of sex information not only improves brain age estimation performance but also does so with fewer trainable parameters. The reduction in parameters is due to the absence of the sex classifier required for multitask learning. Most importantly, the proposed SA-AVAE framework provides balanced performance across both male and female groups, demonstrating that the framework not only enhances brain age estimation but also improves robustness.

\begin{table}[]

\caption{\updatedText{Comparison of models with and without sex information incorporation in the proposed adversarial variational autoencoder (AVAE) performance, along with the number of parameters, evaluated on different metrics for overall, male, and female groups.}}
\centering
\begin{tabular}{ccll}
\hline
\rowcolor[HTML]{EFEFEF} 
\multicolumn{1}{l}{\cellcolor[HTML]{EFEFEF}\textbf{Method}} &
  \multicolumn{1}{l}{\cellcolor[HTML]{EFEFEF}\textbf{\begin{tabular}[c]{@{}l@{}}No. of \\ Parameters\end{tabular}}} &
  \textbf{Group} &
  \multicolumn{1}{c}{\cellcolor[HTML]{EFEFEF}\textbf{MAE}} \\ \hline
                          &                          & Female  & 2.813 ± 1.626 \\ \cline{3-4} 
                          &                          & Male    & 2.979 ± 1.594 \\ \cline{3-4} 
\multirow{-3}{*}{AVAE}    & \multirow{-3}{*}{2.70 M} & Overall & 2.921 ± 1.608 \\ \hline
                          &                          & Female  & 2.620 ± 1.528 \\ \cline{3-4} 
                          &                          & Male    & 2.863 ± 1.508 \\ \cline{3-4} 
\multirow{-3}{*}{M-AVAE \cite{usman2024multi, usman2024advancing}}  & \multirow{-3}{*}{3.15 M} & Overall & 2.779 ± 1.520 \\ \hline
                          &                          & Female  & 2.713 ± 1.363 \\ \cline{3-4} 
                          &                          & Male    & 2.727 ± 1.345 \\ \cline{3-4} 
\multirow{-3}{*}{SA-AVAE} & \multirow{-3}{*}{2.70 M} & Overall & 2.722 ± 1.351 \\ \hline
\end{tabular}
\label{sex_info_tab}
\end{table}

\subsection{\updatedText{Limitations and Future Work}}
\updatedText{The proposed SA-AVAE framework leverages two neuroimaging modalities, structural MRI (sMRI) and functional MRI (fMRI), for the accurate estimation of biological brain age. While the integration of both modalities enhances the model's performance and improves the accuracy of brain age estimation, it also introduces certain limitations. One of the primary challenges is the framework's sensitivity to missing modalities. As discussed in Section \ref{multimodal_analysis_section}, the performance of the SA-AVAE model significantly deteriorates when either modality is unavailable. This makes the model highly dependent on the availability of both sMRI and fMRI data, which can be a significant limitation in practical clinical scenarios where one of the modalities may be missing or of poor quality. Another major limitation of this study is the exclusive use of data from healthy control (HC) subjects. While the results obtained on this cohort are promising, the real-world applicability of the framework for early detection of neurodegenerative diseases, such as Alzheimer's Disease (AD), remains uncertain. To more effectively evaluate the performance and generalizability of the proposed framework, it is crucial to test it on patients with various neurological conditions, including AD and Parkinson's Disease (PD). This will allow for a more thorough assessment of how the model performs in distinguishing biological brain age across different patient populations, which is essential for its potential clinical use.}

\updatedText{Future work includes several key improvements that are necessary for the framework to achieve its full potential. One such area is enhancing the model's robustness so that it can perform comparably well with either sMRI or fMRI data alone, thereby reducing the performance gap between multimodal and unimodal approaches. This would increase the model’s flexibility, enabling its deployment in situations where only a single modality is available or when multimodal data collection is not feasible. Additionally, we plan to conduct a comprehensive evaluation of the SA-AVAE framework using clinical data from patients with various neurological conditions. This will involve real-time testing in clinical settings to assess the feasibility of using the framework for the early diagnosis and monitoring of brain aging and neurodegenerative diseases. These efforts will be critical in demonstrating the clinical utility and robustness of the proposed framework in real-world applications.}

\section{Conclusion}
\label{con}
In this study, we introduced a novel framework for biological brain age estimation, leveraging the complementary information from structural magnetic resonance imaging (sMRI) and functional magnetic resonance imaging (fMRI) data. Our proposed Sex-Aware Adversarial Variational Autoencoder (SA-AVAE) combines adversarial and variational learning techniques to effectively disentangle latent features from both modalities. By decomposing the latent space into modality-specific and shared codes, our model captures both the unique and common aspects of brain aging, while also accounting for sex-specific aging patterns, further enhancing its performance. The results of our experiments, evaluated on the OpenBHB dataset, demonstrate that SA-AVAE outperforms existing state-of-the-art methods, showing significant robustness across various age groups. This highlights the potential of the framework for real-time clinical applications, particularly in the early detection and monitoring of neurodegenerative diseases. Future work will focus on enhancing the robustness of the framework to ensure effective performance with either modality independently. Additionally, we aim to conduct real-time evaluations using clinical data to assess the practical applicability of the SA-AVAE model in clinical settings.



\end{document}